\documentclass[conference]{IEEEtran}
\IEEEoverridecommandlockouts
\usepackage{flushend,cite}
\usepackage{amsmath,amssymb,amsfonts}
\usepackage{algorithmic}
\usepackage{graphicx}
\usepackage{textcomp}
\usepackage{xcolor}
\usepackage{booktabs}

\usepackage{subfigure}
\usepackage[hyphens]{url}
\usepackage{hyperref}
\usepackage{booktabs}
\usepackage{placeins}
\usepackage{float}
\restylefloat{figure}

\usepackage{sidecap}

\usepackage[hyphens]{url}
\usepackage{hyperref}
\usepackage{booktabs}
\usepackage{tikz}
\usetikzlibrary{positioning}
\usetikzlibrary{shapes.geometric}

\usepackage{balance}

\def\BibTeX{{\rm B\kern-.05em{\sc i\kern-.025em b}\kern-.08em
    T\kern-.1667em\lower.7ex\hbox{E}\kern-.125emX}}
\begin{document}

\title{Multitemporal analysis in Google Earth
Engine\\ for detecting urban changes  using optical data\\ and machine learning algorithms
}
\makeatletter
\newcommand{\newlineauthors}{%
  \end{@IEEEauthorhalign}\hfill\mbox{}\par
  \mbox{}\hfill\begin{@IEEEauthorhalign}
}
\makeatother

\author{\IEEEauthorblockN{1\textsuperscript{nd} Mariapia Rita Iandolo}
\IEEEauthorblockA{\textit{Engineering Department} \\
\textit{University of Sannio}\\
Benevento, Italy \\
m.iandolo1$@$studenti.unisannio.it}
\and

\IEEEauthorblockN{2\textsuperscript{st} Francesca Razzano}
\IEEEauthorblockA{\textit{Engineering Department} \\
\textit{University of Sannio}\\
Benevento, Italy  \\
f.razzano3$@$studenti.unisannio.it}
\and
\IEEEauthorblockN{3\textsuperscript{rd} Chiara Zarro}
\IEEEauthorblockA{\textit{Aerospace Department} \\
\textit{Intelligentia srl}\\
Benevento, Italy \\ chiara.zarro@intelligentia.it
}
\and
\newlineauthors

%\IEEEauthorblockN{4\textsuperscript{th} Alessandro Sebastianelli}
%\IEEEauthorblockA{\textit{$\phi$-lab} \\
%\textit{European Space Agency}\\
%Frascati, Italy  \\
%Alessandro.Sebastianelli@esa.int}
\IEEEauthorblockN{4\textsuperscript{th} G. S. Yogesh}
\IEEEauthorblockA{\textit{Engineering Department} \\
\textit{East Point College}\\
Bangalore, India  \\
gs.yogesh@eastpoint.ac.in}
\and
\IEEEauthorblockN{5\textsuperscript{th} Silvia Liberata Ullo}
\IEEEauthorblockA{\textit{Engineering Department} \\
\textit{University of Sannio}\\
Benevento, Italy \\
ullo@unisannio.it}
}

\maketitle

\begin{abstract}
The aim of this work is to perform a multitemporal analysis using the Google Earth Engine (GEE) platform for the detection of changes in urban areas using optical data and specific machine learning (ML) algorithms. 
%In particular, after dealing with the origins and foundations of Remote Sensing, the concepts related to classification and change detection will be addressed. Specifically, in reference to the application case, we will acquire data of a particular region of interest of which the classification will be made through Machine Learning algorithms implemented on the Google Earth Engine platform. 
As a case study, Cairo City has been identified, in Egypt country, as one of the five most populous megacities of the last decade in the world. 
%The purpose of the case study is to verify, given a time frame, whether or not there have been changes in the urban area considered. 
%After performing the 
Classification and change detection analysis of the region of interest (ROI) have been carried out 
from July 2013 to July 2021.
%, it was shown that, as expected, From the data collected, it has been found that there have been real changes in the urban area of Cairo. An accurate analysis of all the results obtained was also reported, or the case where a non-urban area remains unchanged or becomes urban and where an urban area remains urban or becomes non-urban. Demonstrated that for each of these four cases there are acquired data that attest to its veracity we can conclude that the objective of the elaborate has been reached with success.
Results demonstrate the validity of the proposed method in identifying changed and unchanged urban areas over the selected period. Furthermore, this work aims to evidence the growing significance of GEE as an efficient cloud-based solution for managing large quantities of satellite data. 
\end{abstract}

\begin{IEEEkeywords}
Optical and SAR data classification, Machine Learning algorithms, change detection, Google Earth Engine, Earth Observation, Landsat-8.
\end{IEEEkeywords}

\section{Introduction}
\noindent 
Nowadays good urban planning cannot be based only on classical methods, but it is important to take advantage of the advanced techniques which are catching on the remote sensing (RS) field. Among them, ML  and Deep Learning (DL) techniques  are increasingly playing a crucial role in handling huge amounts of satellite data and enhancing results in terms of classification analysis, land use monitoring, and natural phenomena detection
\cite{param}, \cite{unnikrishnan2019deep}, %\cite{mishra2020land}, 
\cite{kadavi2018land}, just to give some examples.  

\noindent Measuring and understanding the nature and entity of changes affecting urban and non-urban territories are crucial in order to determine their future expansion and impact in terms of environmental and economic issues.
For this reason, extensive studies and research have been carried out on detecting urban changes and on using their results as valuable information to support Governments and Municipalities in taking better decisions. In \cite{10150757}, change detection is applied to Newly Constructed Areas (NCA) as the first step in the 
development monitoring of urban areas, through a ML approach. 
%In this field,
%Works of Classification and Detection of Changes in Urban and Non-Mourbal Areas have been carried out in works such as \cite{primocite}, \cite{secondocite}, %\cite{detectioncoroba}, 
%to name a few.
Various spectral indices for a  rapid and accurate built land classification are proposed in \cite{primocite}. Their  performance is examined and compared in the classification and detection of land changes when Landsat-images7 ETM+ (Enhanced Thematic Mapper Plus) and Landsat-8 OLI/TIRS (Operational Land Imager/Thermal Infrared Sensor) are used as satellite images. In \cite{secondocite}, authors  deal with the change detection of the urban area of Bauchi which is one of the cities in the northeastern part of Nigeria that has witnessed a huge expansion due to rapid urbanization. In particular, they compare three change detection algorithms, supervised, unsupervised, and post-classification comparison, the latter for achieving a "from-to" evaluation, and demonstrate that the supervised classification produces the best  results in terms of  overall precision in the reference years.  %of 93.5 \% and 89.7 \% for 2003 and 2013 respectively.  

In our paper, we aim to perform a multitemporal analysis of optical satellite data for the detection of urban changes when specific ML algorithms are used and the GEE platform is employed. GEE has gained great attention for its offer of built-in solutions above all for what concerns the ML and DL algorithms. Therefore, our work aims also to demonstrate the growing significance of GEE as an efficient cloud-based tool for processing huge amounts of satellite data producing valuable results and useful information in the field of RS and urban planning. The pivotal phase of our work remains the application of change detection to remote sensing images belonging to different time periods (almost ten years are considered) for the identification and discrimination of urban variations. Change detection can allow keeping track of occurred modifications  otherwise not detectable on official documentation or through sample surveys.
%The purpose of this work is to show one of the many functions and applications of remote sensing, specifically that one according to which acquiring images from satellite, using a geospatial processing service called Google Earth Engine and, Through Machine Learning algorithms, on time distant from each other, changes can be identified and not, in urban area due to the expansion of the latter or its invariance.
Cairo City shown in Figure \ref{cairo} has been chosen as the case study  and supervised ML algorithms on GEE have been selected.

\section{GEE and Data sources}
\noindent In order to perform the supervised classification process and the subsequent change detection, GEE was used with the  supervised ML techniques made available on the platform.  The satellite images were downloaded still through GEE, and imagery of the   \href{https://landsat.gsfc.nasa.gov/satellites/landsat-8/landsat-8-mission-details/}{Landsat-8}  mission was considered, a mission born thanks to the   collaboration between NASA and the U.S. Geological Survey.

\subsection{Google Earth Engine}
\noindent 
\href{https://earthengine.google.com/#intro}{GEE} is a powerful web platform for the cloud-based processing of large-scale remote sensing data. It brings together many world satellite images and makes them available online for scientists, independent researchers, and nations who want to use them and to detect changes, monitor the atmosphere, map trends, and quantify the differences in the Earth’s surface. The advantage lies in its remarkable calculation speed since the processing is outsourced to Google servers. Application programming interfaces (APIs) are  available in JavaScript and Python with code editor features designed to make developing complex geospatial workflows quick and easy.
%\cite{gee}
There are different ML techniques that can be used in GEE, such as  
\href{https://developers.google.com/earth-engine/guides/machine-learning}{Supervised and Unsupervised Classification}.
%\cite{ML}.
\\
In particular, in our investigation, we made use of the  package handling supervised classification and we  chose the CART 
"ee.Classifier.smileCart()" tool. The CART algorithm was first published by Leo Breiman in 1984 \cite{Breiman}, and it is a decision tree model based on if-else rules, able to predict outcome values based on other quantities. The  \textit{SmileCart} ML algorithm was therefore applied to Landsat-8 data related to the urban agglomeration of Cairo City, chosen as ROI.
\subsection{Landsat-8}
\noindent \href{https://landsat.gsfc.nasa.gov/satellites/landsat-8/landsat-8-mission-details/}{Landsat-8}
%\ref{l8} (formerly the Landsat Data Continuity Mission, or LDCM) 
was launched on 
%an Atlas-V rocket from Vandenberg Air Force Base, California on 
February 11, 2013, and it is  collecting about 740 scenes per day on the Worldwide Reference System-2 (WRS-2). With a revisit cycle of 16 days, it carries on board the Operational Land Imager (OLI) and Thermal Infrared Sensor (TIRS) instruments, featuring  11 bands with the addition of a first ultra-blue band (0.43-0.45 micrometers), nine bands useful for displaying clouds (1.36-1-38 micrometers) and the thermal band split into two separate bands TIR1 (10.60-11.19 micrometers) and TIR2 (11.50-12.51 micrometers) with 100 meters resolution. 
%Landsat-8 instruments represent a technological advance in the evolution of satellite systems. OLI improves past Landsat sensors using a more technical approach. In addition, Landsat-8 collects data for the spectral bands of visible infrared, near infrared and short wave, as well as a panchromatic band. 
%It has a design life of five years.
%Landsat-8 orbits the Earth in an almost polar helio-synchronous orbit at an altitude of 705 km (438 mi), tilted by 98.2 degrees, and completes an Earth orbit every 99 minutes. The satellite has a repeat cycle of 16 days.
%Landsat 8 captures . The dimensions of a Landsat-8 scene are 185 km x 180 km. 
%The data products created by the Landsat 8 OLI/TIRS scenes are available for download from EarthExplorer, Glovis and LandLook Viewer. 
%\cite{Landsat8888}
%\begin{figure}[!ht]
%	\centering
 %   \includegraphics[width= 0.47\textwidth]{LDCM_inorbit_low_81fddc71-bc2f-4699-affc-20ed77e83f33-prv.jpg}
%	\caption{ Landsat 8 (formerly the Landsat Data Continuity Mission, or LDCM)}
%	\label{l8}
%\end{figure}
\section{Region of Interest}
\noindent The area chosen for the specific case study is 
\href{https://earth.esa.int/web/earth-watching/image-of-the-week/content/-/article/cairo-egypt-sentinel/}{Cairo City}, 
the capital of Egypt.
%, which was founded on the Nile. 
It has about 18 million inhabitants living inside the urban area (spread over 453 $km^2$) and about 20.4 million residents in the adjacent metropolitan area, creating the mega metropolis Cairota, Greater Cairo, which is the most populous African city after the Nigerian city Lagos. 
%\cite{cairo}
%According to a recent study Cairo is among the most populous future megacities of 2030.   cite{megalopolis}
This area has been chosen since the purpose of the proposed method is to offer a way of analysis over time to 
verify whether or not there have been changes in a certain urban area  and if at some points there is an extension of the urban agglomeration or if instead, it has remained unchanged, if some urban areas have become non-urban and vice versa, 
or have simply remained not urban. 
\begin{figure}[ht!]
	\centering
\includegraphics[scale=0.25]{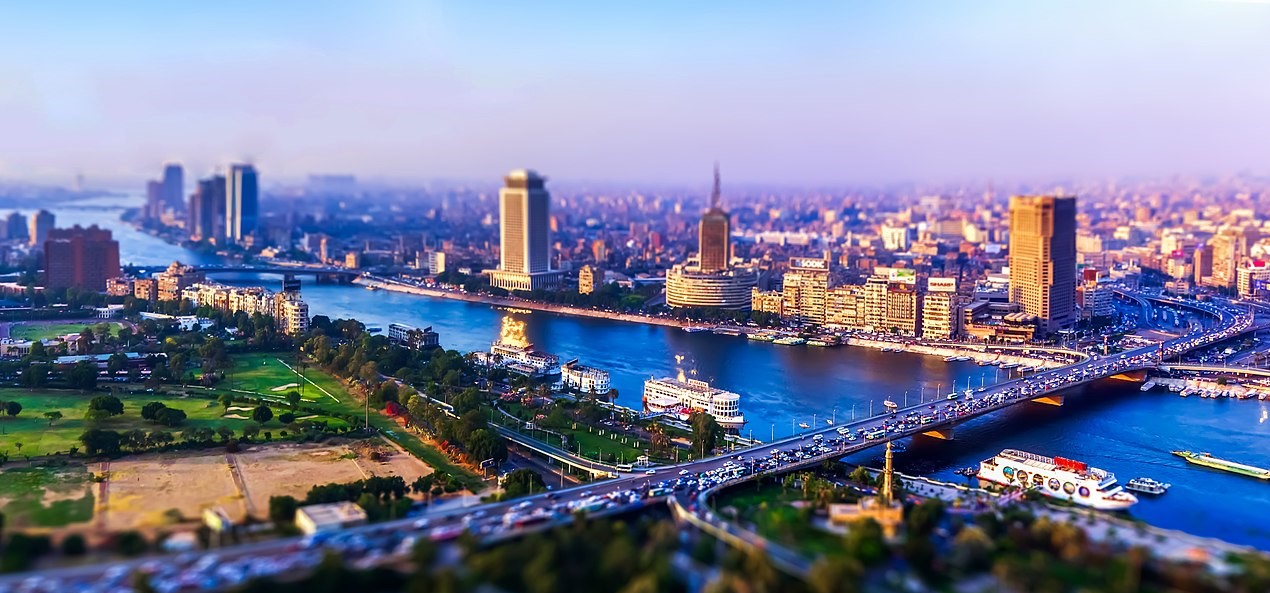}
	\caption{Il \href{https://earth.esa.int/web/earth-watching/image-of-the-week/content/-/article/cairo-egypt-sentinel/}{Cairo City}}
	\label{cairo}
\end{figure}

\section{Methodology}
%\vspace{-1cm}
\noindent The following procedure has been applied for the classification and change detection analysis of the chosen area.
\subsection{Classification}
\noindent For classification purposes, useful guidelines  can be found at \cite{website1} and \cite{classificazione}, and the main steps  considered in GEE are reported in the Block Diagram of Figure \ref{class}.\\ 
\begin{figure}[ht!]
	\centering
\includegraphics[scale=0.45]{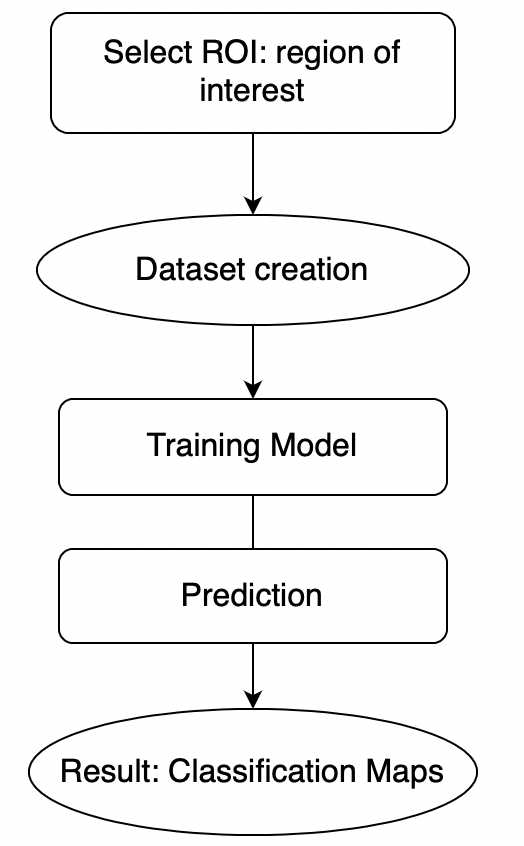}
	\caption{Block Diagram  for Classification steps in GEE \href{}{}}
	\label{class}
\end{figure}
\\
In our case, the following description summarizes what done:
\begin{itemize}
\item the \textit{Roi}, Cairo City, with coordinates [31.206,30.248], has been selected;
\item an image Landsat-8 has been chosen for the oldest date of the chosen time period, that is 2013, in the month of July to avoid problems related to the possible presence of clouds; from this image, a dataset is created for the training model. A feature collection (a collection of GeoJSON objects) has been created, assigning each sample a label, in particular: \textit{0} for samples of the non-urban class, \textit{1} for samples of the urban class. This phase in the block diagram is called \textit{Dataset creation};
\item after instantiating the selected classifier (the \textit{SmileCart} classifier), and setting its parameters, the classifier has been trained with the training data. This phase in the block diagram is called the \textit{Training Model};
\item the trained model has been used for the prediction of all other images (so from 2013 to 2021). This phase is called \textit{Prediction} after which the results are the \textit{Classification Maps}.
\end{itemize}

\subsection{Change Detection}
\noindent Once the classification data for each year have been acquired and assessed, we proceeded with the change detection step by taking into account the oldest classification (2013) and the most recent one (2021), to detect in a fairly large period of time changes or invariances, and using intermediate years for additional checking and comparisons.\\ 
Four possible results have been considered: 
\begin{enumerate}
 \item the case where there has actually been an expansion of the urban agglomeration, this result in the acquired image corresponds to the red color;
 
 \item the case in which there was no expansion of the urban agglomeration that just remained the same as it was on the older date considered, this result in the acquired image corresponds to the color purple;
 
\item the case where an originally urban environment became non-urban, this result in the acquired image corresponds to the blue color; 

\item and finally the case where an originally non-urban environment is still not urban and undergoes no change, this result in the acquired image corresponds to the  green color.
\end{enumerate}  

%\noindent Results of change detection refer to possible modifications that could take place from July 2013 to July 2021. The purpose is to verify  that the four possible results, as described above, really occurred and corresponded to the truth.
%This analysis is possible, as we have acquired the classification of the same area even in the years between 2013 and 2021 (always in the month of July), even considering different time frames, for example from 2015 to 2018.
%However, for change detection, we focused, as already mentioned, only on the time frame that goes from July 2013 to July 2021.

\section{Results and discussion}
\noindent 
We used Landsat-8 images with almost 10 years of difference, for which qualitative evaluation could allow us to establish the occurred variations, if any. From change detection images it has been possible, as discussed ahead, to prove the proposed methodology. Moreover, further analysis has been applied to smaller areas when some additional investigation was needed.\\
We started from the simplest case, which is the one that refers to the non-change of one of the non-urban areas around the megalopolis considered: the river Nile and the land around it.
From  Figure  \ref{nilo} (a), referring to the year 2013, acquired through Landsat-8, you can easily distinguish the river Nile. If we  consider the same image but acquired in the year 2021 (Figure \ref{nilo} (b)), through visual inspection,  any significant change related to the watercourse is noticed. As a result of the performed change detection the area is mainly purple with some red spots in other places that will be further analyzed. In the area surrounding the Nile, it is possible to verify how the overall method and the change detection image were able to convey  a clear situation of what happened.  
%that you go to perform, as mentioned above in the paragraph of the application case, the stream should turn green.

\begin{figure}[ht!]
	\centering
\resizebox{.48\textwidth}{!}{%
\includegraphics[height=3cm]{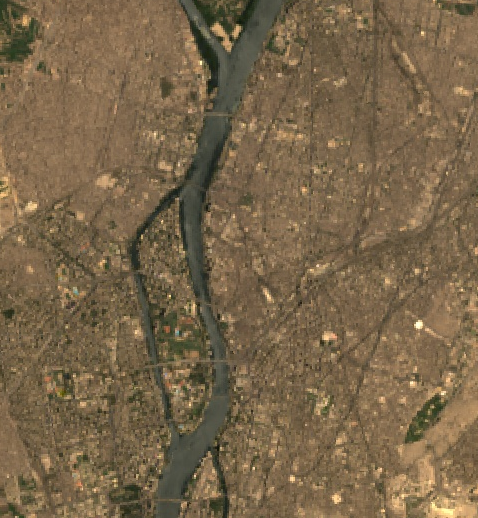}%
\quad
\includegraphics[height=3cm]{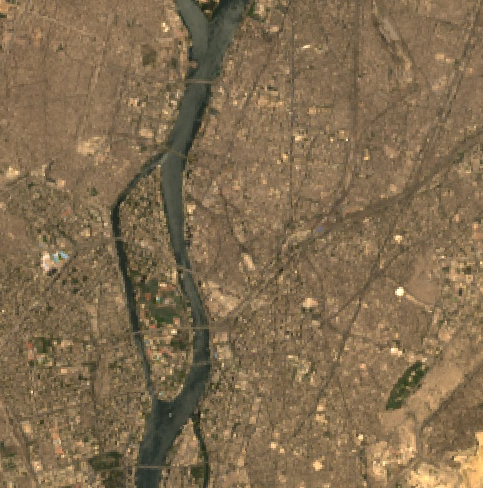}%
\quad
\includegraphics[height=3cm]{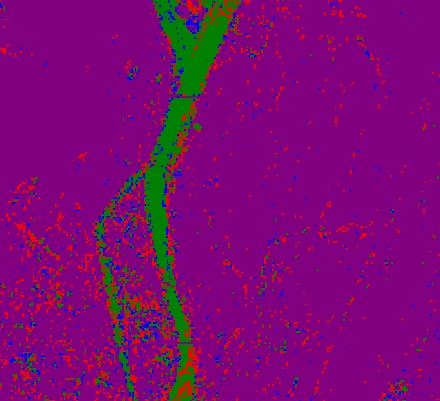}%
}
\caption{(a) Satellite image for 2013:  Nile river. (b) Satellite image for 2021 of the same river. (c)  Change detection image: the  Nile River and its surrounding area have not changed}
	\label{nilo}
\end{figure}

As a result,  we got that there were actually no changes concerning this area, from non-urban it continued to remain non-urban.\\
\noindent 
\noindent We then considered the case in which one of the many areas surrounding the megalopolis from non-urban became urban. Namely, a real extension of the urban agglomeration occurred. 
The city of Asim was considered again by taking into account first the image acquired by the satellite in 2013 (Figure \ref{asim}) (a) and then the image of the same area  in 2021 (Figure \ref{asim}) (b). 
What you can notice through visual inspection is that the urban agglomeration of Asim has expanded. So the non-urban area around it should turn red as mentioned above. Once checked with the proposed method and the change detection output,  from Figure \ref{asim} (c) the city of Asim has effectively expanded:  the whole area surrounding its borders has been colored  red. This means that the surrounding non-urban area has undergone a change, turning into an urban area.

\begin{figure}[ht!]
	\centering
\resizebox{.48\textwidth}{!}{%
\includegraphics[height=3cm]{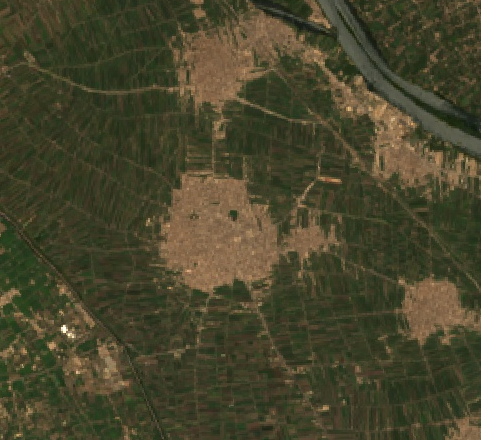}%
\quad
\includegraphics[height=3cm]{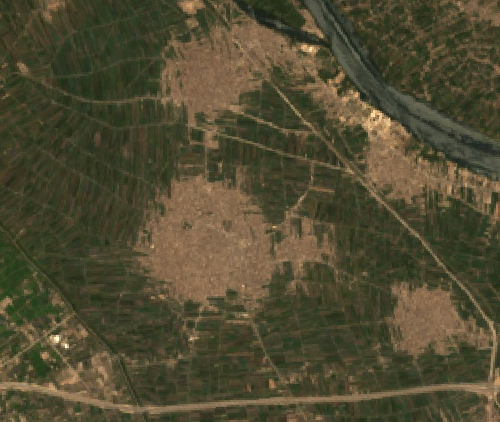}%
\quad
\includegraphics[height=3cm]{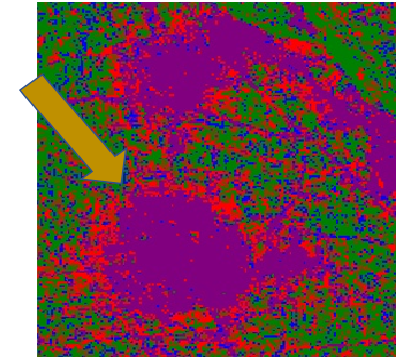}%
}
\caption{(a) Satellite image for 2013: the city of Asim. (b) Satellite image for 2021 of the same Asim city. (c)  Change detection image: Asim’s city has expanded (red circular crown).}
	\label{asim}
\end{figure}

\noindent We have considered as the penultimate case  an urban area surrounding the Cairota megalopolis that did not undergo any change, that is urban continued to be urban.
We chose to analyze the city of Tanta, again with an acquisition first  in 2013 and then in 2021 (Figure \ref{tanta} (a) e (b)). From a visual inspection of the satellite images, it can  be said that between 2013 and 2021 there were no major changes and that the area of Tanta remained unchanged. As expected, the proposed method and the final change detection produced a final image colored all purple, according to what was said before.  
In conclusion, as seen from Figure \ref{tanta} (c), the city of Tanta  in the chosen time frame has not undergone any change. However, just a few blue and red spots are present and we decided to further investigate the blue ones, representing a new situation among those already analyzed.

\begin{figure}[ht!]
	\centering
\resizebox{.48\textwidth}{!}{%
\includegraphics[height=3cm]{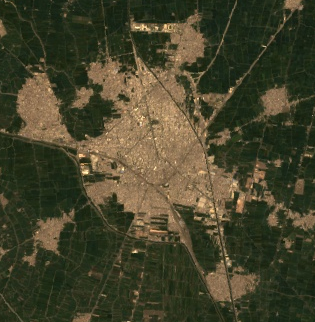}%
\quad
\includegraphics[height=3cm]{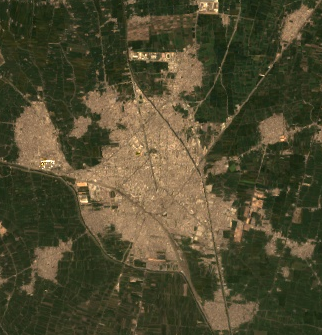}%
\quad
\includegraphics[height=3cm]{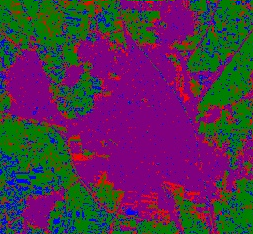}%
}
\caption{(a) Satellite image for 2013: the city of Tanta in its entirety. (b) Satellite image of the whole of Tanta city in 2021. (c) Change detection image: the vast majority remained unchanged with a few blue pixels in the bottom left corner (further investigated).}
	\label{tanta}
\end{figure}

%\begin{figure}[ht!]
%	\centering
% \subfigure[]{
	%\includegraphics[scale=0.36]{immagini/tanta2013.PNG}}%
% \subfigure[]{
% \includegraphics[scale=0.33]{immagini/tanta2021.PNG}}%
% \subfigure[]{
 %\includegraphics[scale=0.43]{immagini/tantachange.PNG}}%
%	\caption{(a) Satellite image for 2013: the city of Tanta. (b) Satellite image for 2021: the city of Tanta. (c) Image taken after Change Detection: Tanta’s city remained unchanged}
%	\label{tanta}
%\end{figure}\\

\noindent In the final case, consideration is given to an urban agglomeration that has become  non-urban within the chosen time frame. We decided to further analyze the portion of territory that in the previous Figure was showing some blue pixels. The zoomed area of this area belonging to the city of Tanta was taken into consideration; acquiring the satellite image in 2013 (Figure \ref{area3} (a)) we could see that the area indicated within the yellow oval in the figure is remotely detected as an urban environment. The image for 2021 (Figure \ref{area3} (b)) cannot be considered the same, since it is remotely detected by the satellite as a non-urban environment. So after performing the change detection, you expect that the area of interest is colored blue.
From Figure \ref{area3} (c) it can be noted that the area of interest has undergone a change, just what was expected. 
The urban area became non-urban and for this reason, it is colored blue. With this last analysis, the case study ends.

\begin{figure}[ht!]
	\centering
\resizebox{.48\textwidth}{!}{%
\includegraphics[height=3cm]{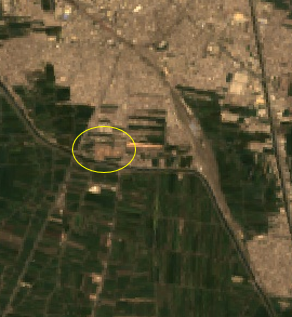}%
\quad
\includegraphics[height=3cm]{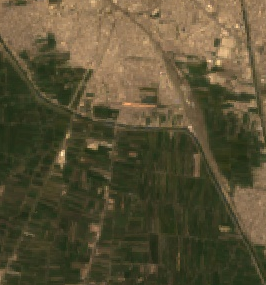}%
\quad
\includegraphics[height=3cm]{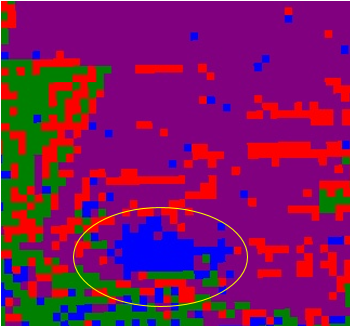}%
}
\caption{(a) Satellite image for 2013: a specific part of Tanta's city (in the yellow oval) is analyzed. (b) Satellite image for 2021: the same part of Tanta's city is considered. (c) Zoomed image of the part in the yellow oval after that change detection was applied: the fraction of Tanta's city changed from urban to non-urban appears in blue.}
	\label{area3}
\end{figure}
%\vspace{1.8cm}
%\columnbreak
\section{Conclusions}
%\noindent 
With this study, we have presented a 
straightforward procedure to verify  the existence of possible changes in the urban area of Cairo and how these changes can be kept under control to monitor urbanization and other modifications (or their absence) over about ten years of observations. From the results obtained, we can say that the performed analysis  was successful, because, point by point, in each case considered (all four cases) we have obtained conclusions  that prove the veracity of this procedure. 
It is worth highlighting that all processing including the ML ones has been carried out inside the GEE platform, offering also the advantage of cloud-based tools and services. Moreover, 
%in the years undergone by the chosen territory. 
%It is necessary to reiterate that 
such an analysis is a small piece of a much larger project. Future work is to extend the analysis to other parts of the world and make further comparisons to assess the proposed methodology in detail. 
%With this application case we have manipulated satellite data recognizing in Google Earth Engine a valuable tool for data processing. Specifically once you have selected a zone on this platform (GGE), of this we have made the classification according to the classes chosen by us and finally we have taken advantage of the classification made to perform a Change Detection and verify through the acquired images if actually in the chosen time span there were or not changes in the urban area.

\bibliographystyle{IEEEtran}
\bibliography{ref}

\begin{thebibliography}{1}
\providecommand{\url}[1]{#1}
\csname url@samestyle\endcsname
\providecommand{\newblock}{\relax}
\providecommand{\bibinfo}[2]{#2}
\providecommand{\BIBentrySTDinterwordspacing}{\spaceskip=0pt\relax}
\providecommand{\BIBentryALTinterwordstretchfactor}{4}
\providecommand{\BIBentryALTinterwordspacing}{\spaceskip=\fontdimen2\font plus
\BIBentryALTinterwordstretchfactor\fontdimen3\font minus
  \fontdimen4\font\relax}
\providecommand{\BIBforeignlanguage}[2]{{%
\expandafter\ifx\csname l@#1\endcsname\relax
\typeout{** WARNING: IEEEtran.bst: No hyphenation pattern has been}%
\typeout{** loaded for the language `#1'. Using the pattern for}%
\typeout{** the default language instead.}%
\else
\language=\csname l@#1\endcsname
\fi
#2}}
\providecommand{\BIBdecl}{\relax}
\BIBdecl

\bibitem{param}
G.~B. Rajendran, U.~M. Kumarasamy, C.~Zarro, P.~B. Divakarachari, and S.~L.
  Ullo, ``{Land-Use and Land-Cover Classification Using a Human Group-Based
  Particle Swarm Optimization Algorithm with an LSTM Classifier on Hybrid
  Pre-Processing Remote-Sensing Images },'' \emph{Remote Sensing}, vol.~12,
  no.~24, p. 4135, 2020.

\bibitem{unnikrishnan2019deep}
A.~Unnikrishnan, V.~Sowmya, and K.~Soman, ``Deep learning architectures for
  land cover classification using red and near-infrared satellite images,''
  \emph{Multimedia Tools and Applications}, vol.~78, pp. 18\,379--18\,394,
  2019.

\bibitem{kadavi2018land}
P.~R. Kadavi and C.-W. Lee, ``{Land cover classification analysis of volcanic
  island in Aleutian Arc using an artificial neural network (ANN) and a support
  vector machine (SVM) from Landsat imagery},'' \emph{Geosciences Journal},
  vol.~22, pp. 653--665, 2018.

\bibitem{10150757}
V.~K.~G. Kalaiselvi, J.~Ranjani, S.~Hariharan, V.~S. Venu, D.~V. K, and P.~P.
  Nandana~K.M, ``{Efficient Change Map Detection from Imagery Data using
  Machine Learning Approach},'' in \emph{2023 International Conference on
  Distributed Computing and Electrical Circuits and Electronics (ICDCECE)},
  2023, pp. 1--4.

\bibitem{primocite}
\BIBentryALTinterwordspacing
R.~C. Estoque and Y.~Murayama, ``{Classification and change detection of
  built-up lands from Landsat-7 ETM+ and Landsat-8 OLI/TIRS imageries: A
  comparative assessment of various spectral indices },'' \emph{Ecological
  indicators}, vol.~56, pp. 205--217, 2015. [Online]. Available:
  \url{https://www.sciencedirect.com/science/article/pii/S1470160X15001673}
\BIBentrySTDinterwordspacing

\bibitem{secondocite}
\BIBentryALTinterwordspacing
K.~M. Kafi, H.~Z.~M. Shafri, and A.~B.~M. Shariff, ``{An analysis of LULC
  change detection using remotely sensed data; A Case study of Bauchi City},''
  \emph{IOP Conference Series: Earth and Environmental Science}, vol.~20,
  no.~1, p. 012056, jun 2014. [Online]. Available:
  \url{https://dx.doi.org/10.1088/1755-1315/20/1/012056}
\BIBentrySTDinterwordspacing

\bibitem{Breiman}
B.~Leo, ``{Classification and Regression Trees},'' \emph{Routledge}, vol. 1st
  ed., 1984.

\bibitem{website1}
G.~E. Engine, ``Supervised classification,'' \url{
  https://developers.google.com/earth-engine/guides/classification}.

\bibitem{classificazione}
J.~T. Point, ``{Algorithm of classification in Machine Learning},''
  \url{https://www.javatpoint.com/classification-algorithm-in-machine-learning}.

\end{thebibliography}
\end{document}